\documentclass[final]{cvpr}

\usepackage{times}
\usepackage{epsfig}
\usepackage{graphicx}
\usepackage{amsmath}
\usepackage{amssymb}
\usepackage{multirow}
\usepackage{color}
\usepackage{colortbl}
\usepackage[colorlinks,linkcolor=blue,breaklinks=true,bookmarks=false]{hyperref}

\def\confYear{CVPR 2021}
\newcommand*{\affaddr}[1]{#1}
\newcommand*{\affmark}[1][*]{\textsuperscript{#1}}

\begin{document}

%%%%%%%%% TITLE
\title{Practical Wide-Angle Portraits Correction with Deep Structured Models}

\author{
    Jing Tan\affmark[1]\thanks {Equal contribution} \quad 
    Shan Zhao\affmark[1]\footnotemark[1] \quad 
    Pengfei Xiong\affmark[2]\footnotemark[1] \quad  
    Jiangyu Liu\affmark[1] \quad 
    Haoqiang Fan\affmark[1] \quad  
    Shuaicheng Liu\affmark[3,1] \thanks {Corresponding author}\\\\

\affaddr{\affmark[1]Megvii Research} \quad
\affaddr{\affmark[2]Tencent}\\
\affaddr{\affmark[3]University of Electronic Science and Technology of China}\\

\tt\small \{tanjing, zhaoshan, liujiangyu, fhq\}@megvii.com \\ \tt\small xiongpengfei2019@gmail.com, liushuaicheng@uestc.edu.cn \\
\tt\small \url{https://github.com/TanJing94/Deep_Portraits_Correction}
}
\maketitle

%%%%%%%%% ABSTRACT
\begin{abstract}
% 现存的问题：1、人像和背景是不同的变形；
% 之前方法的问题：1、只能解决其中某一个变形，2、依赖于标定参数；3、需要复杂的迭代；
% 创新点：1、第一个深度学习方法；2、一个级联的方法分别解决背景和人像，并实现平滑过渡；3、提出了新的metric；4、提供了一个diversity的数据集；5、效果比之前的好，尤其是过渡区域；
Wide-angle portraits often enjoy expanded views. However, they contain perspective distortions, especially noticeable when capturing group portrait photos, where the background is skewed and faces are stretched. This paper introduces the first deep learning based approach to remove such artifacts from freely-shot photos. Specifically, given a wide-angle portrait as input, we build a cascaded network consisting of a LineNet, a ShapeNet, and a transition module (TM), which corrects perspective distortions on the background, adapts to the stereographic projection on facial regions, and achieves smooth transitions between these two projections, accordingly. To train our network, we build the first perspective portrait dataset with a large diversity in identities, scenes and camera modules. For the quantitative evaluation, we introduce two novel metrics, line consistency and face congruence. Compared to the previous state-of-the-art approach, our method does not require camera distortion parameters. We demonstrate that our approach significantly outperforms the previous state-of-the-art approach both qualitatively and quantitatively.
\end{abstract}

%%%%%%%%% BODY TEXT
\section{Introduction}
\begin{figure}[t]
	\begin{center}
		\includegraphics[width=1.0\linewidth]{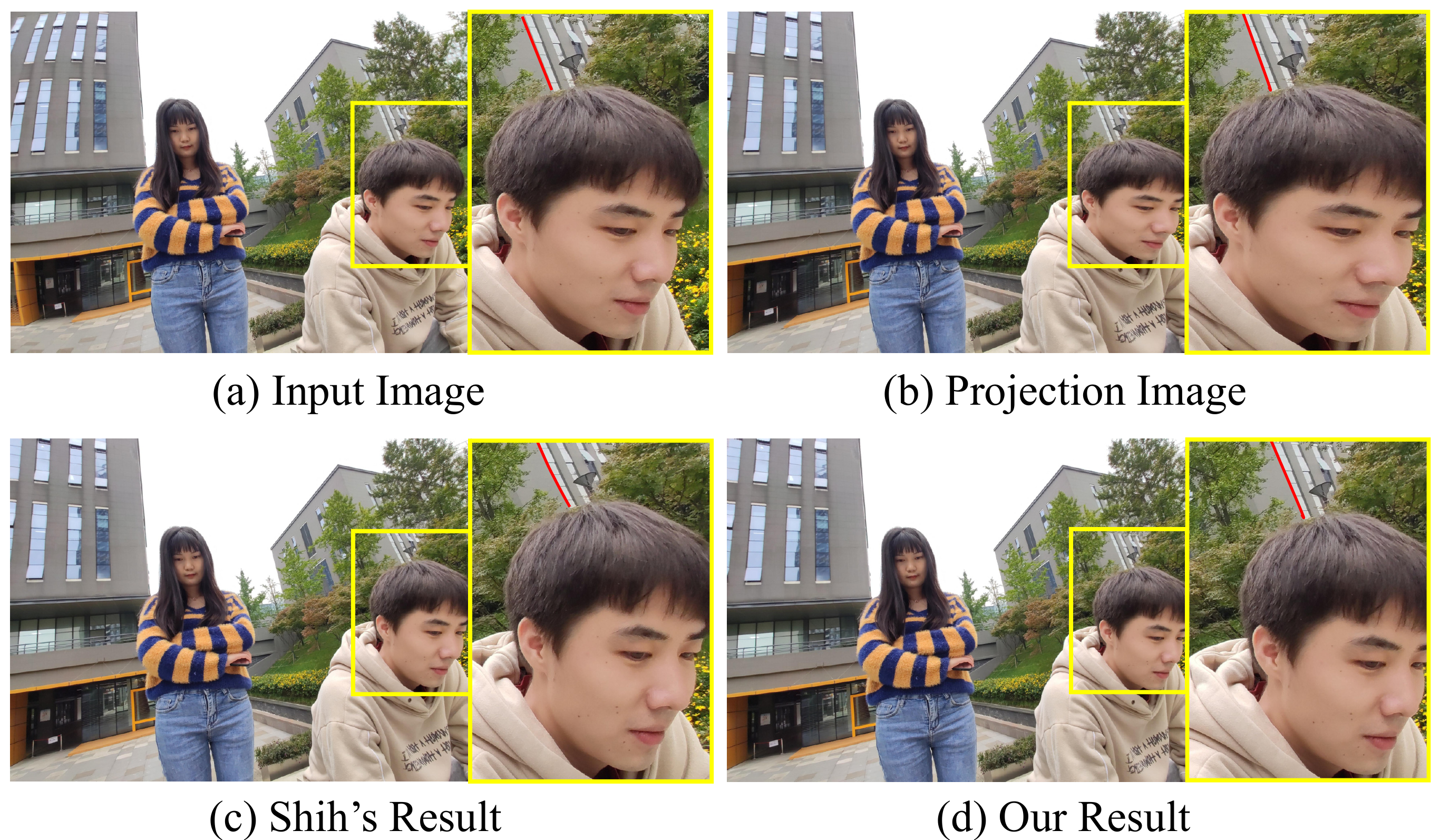}
	\end{center}
	\vspace{-3mm}
	\caption{Examples of distorted and corrected photographs. (a) the original distortion image with curved background and distorted faces. (b) projection image with straight lines. (c) result by Shih~\emph{et al.}~\cite{shih2019distortion}, and (d) result of the proposed deep learning method. Both the background and faces are corrected in (d).}
	\vspace{-3mm}
	\label{fig:example}
\end{figure}

% 随着手机广角摄像头的不断普及，广角图像越来越受欢迎，因为其更大的视野，和更丰富的层次关系；但是广角也带来了明显的畸变。由于光心和图像的距离不同，以及纹理表现不同，背景和人脸往往呈现不同的扭曲。如图1所示，人像的扭曲很小，而边缘的扭曲更大
With the popularity of wide-angle cameras on smartphones, photographers can take pictures with broad vision.
%shoot broad vision and rich layers of scenic background. 
However, a wider field-of-view often introduces a stronger perspective distortion. All wide-angle cameras suffer from distortion artifacts that stretch and twist buildings, road ridges and faces, as shown in Fig.~\ref{fig:example} (a). 

% 图像矫正是一个很难的问题。传统的方法都是直接基于标定参数把图像投影到平面，但这样会导致人脸区域也被拉伸。其他的方法则尽量保持主体的不变形，来优化背景区域。
There are relatively few works targeting on the perspective distortion correction in portrait photography\cite{fried2016perspective,Cooper2012}. Previous methods apply perspective undistortion using camera calibrated distortion parameters~\cite{pavic2006interactive,tehrani2016correcting,carroll2010artistic,du2013changing}, which projects the image onto a plane for undistortion, as shown in Fig.~\ref{fig:example} (b). Compared with Fig.~\ref{fig:example} (a), the lines at the background become straight. Unfortunately, the faces are also projected as a plane, becoming unnaturally wider and asymmetric. It is then evident that background and faces require different types of corrections, to be separately handled with different strategies. As traditional calibration-based methods\cite{Datta2009,Heikkila2000,Chen2004,Geoffroy2018} can only correct distortion in background regions, we need new ways to process faces. 
%Some other methods add constraints onto the distorted subjects in the process of projection, in order to reduce the stretch ratio~\cite{}. However, it is challenging to balance the undistortion of backgrounds and portrait areas and sometimes require large amount of user interactions for good results. 

% 这里有两种写法：
% 1、google提出了一种方法，但是它存在一些问题，我们提出了一个新的深度学习的方法；
% 2、为了解决平衡的问题，我们提出了一种深度学习的方法。之前google也有一个类似的方法，但是XX
Recently, Shih~\emph{et al.}~\cite{shih2019distortion} proposes to deform a mesh which adapts to the stereographic projection~\cite{svardal2003stereographic} on facial regions, and applies perspective projection over the background, enabling different handling of the background and faces. However, a new problem arises, where the smooth transition between faces and background regions is non-trivial. In addition, the method~\cite{shih2019distortion} requires camera distortion parameters as well as the portrait segmentation mask as additional inputs. Fig.~\ref{fig:example} (c) shows the result, where the face in the corner has been over corrected and appear deformed. 

In contrast, our approach does not rely on any prior calibrated parameters, thus being more flexible to various conditioned portraits. Compared to the mesh-based energy minimization~\cite{shih2019distortion}, our deep solution works well in balancing the perspective projection on the background and stereographic projection on the faces, delivering smooth transitions between them. %with few artifacts
%like bending of straight lines and distortion of faces. 
Fig.~\ref{fig:example} (d) shows our result.

%基于此，本文提出了一个深度学习的方法端到端的实现人像的变形。我们的网络由两个部分组成，线矫正网络和人像矫正网络。给定一张畸变图像，首先一个线矫正网络用于生成透视投影的mesh，将该图像投影到平面。其次，投影图像被输入到人像矫正网络进行第二层矫正。为了使得两个网络分别作用于不同的变形，我们设计了一个self-attention模块分别作用在两个网络上。具体而言，两个相似的unet结构被采用从输入的图像中学习出每个像素点的偏移量。self-attention模块作为一个extra的分支，分别学习出变形的感兴趣区域，继而乘在原始输出上强化变形区域。进一步，为了实现两种变形之间的平滑过渡，我们设计了一个trainsion模块，将第一层网络的featmap用于第二层网络。
To this end, we propose a deep structured network to generate a content-aware warping flow field, which both straightens the background lines through perspective undistortion, and adapts to the stereographic projection on facial regions, notably achieving smooth transitions between these two projections. Our cascaded network includes a Line Correction Network (LineNet) and a Portrait Correction Network (ShapeNet). Specifically, given an input image, the LineNet is first applied to produce a flow field to undistort the perspective effects for line correction, where a Line Attention Module (LAM) is introduced to facilitate the localization of lines. Second, the projected image is fed into the ShapeNet for face correction, within which a Face Attention Module (FAM) is introduced for face localization. Furthermore, we design a Transition Module (TM) between LineNet and ShapeNet to ensure smooth transitions. 

%训练我们的模型需要一个大数据集。据我们所知，之前并没有这样一个数据集。因此，我们重建了一个数据集，人工交互XXX；Furthermore，为了定量的描述我们模型的精度，我们设计了两个指标，分别来衡量背景线条和人像的真实程度。
As there is no proper dataset readily available for training, we build a high-quality wide-angle portrait dataset. Specifically, we capture portrait photos by smartphones with various wide-angle lenses and then interactively correct them with a specially designed content-aware mesh warping tool, yielding $5,854$ pairs of input and output images for training. Moreover, for quantitative evaluations, we introduce two novel metrics, Line Straightness Metric (LineAcc) and Shape Congruence Metric (ShapeAcc) to evaluate the line straightness and face correctness accordingly. Previously, evaluation can only be made qualitatively. 

Experimental results show that our approach can correct distortions in wide-angle portraits. Compared with calibration-based opponents, our method can rectify the faces faithfully without camera parameters. Compared with Shih's method~\cite{shih2019distortion}, our method is calibration-free, and achieves good transitions between background and face regions. Both qualitative and quantitative evaluations are provided to validate the effectiveness of our method. Our main contributions are:

%我们的创新点可以归纳为：
%1、我们是第一个深度学习的图像畸变算法；我们提出了一个深度模型；
%2、我们的算法不依赖于相机参数，据有更好的普适性。
%3、attention和coord的模块使得泛化能力提升。
%4、我们提供了一个标准数据集，用于图像畸变算法的研究。而且两个新的metric被提出来用于定量测试；
\begin{itemize}
\item The first deep learning based method to automatically remove distortions in wide-angle portraits from unconstrained photographs, 
without camera calibration and distortion parameters,
%where no camera calibration and distortion parameters are required, 
delivering a better universality.  
\item We design a structured network to remove the distortion on background and portraits respectively, achieving smooth transitions
%where smooth transitions can be realized 
between perspective-rectified background and stereographic-rectified faces. 

\item We provide a new perspective portrait dataset for image undistortion with a wide range of subject identities, scenes and camera modules. In addition, two universal metrics are designed for the quantitative evaluation. 
\end{itemize}

\section{Related Works}
\subsection{Image Distortions}
Image distortions are often introduced when projecting a 3D scene to a 2D image plane through a limited Field-of-View (FOV)~\cite{zorin1995correction}. The perspective projection often distorts objects that are far away from the camera center~\cite{vishwanath2005pictures}. Mesh-based methods have been attempted with user constraints, e.g., dominant straight lines and vanishing points, to cope with potential undesired mesh deformations~\cite{carroll2009optimizing,kanamori2011local}. In this work, our method is calibration-free, which not only rectifies background perspective distortions, but also corrects the face distortions with a deep neural network. 

\subsection{Content-Aware Warping}
Mesh-based content-aware warping have been widely applied in image and video manipulations, including image stitching~\cite{zaragoza2013projective,chang2014shape}, video stitching~\cite{guo2016joint,lin2016seamless,jiang2015}, panorama rectangling~\cite{he2013rectangling}, content-aware rotation~\cite{he2013content}, perspective manipulation~\cite{carroll2010artistic}, image retargeting~\cite{wang2008optimized}, video retargeting~\cite{wang2009motion,wang2010motion}, stereoscopic editing~\cite{chang2011content}, and video stabilization~\cite{liu2009content,liu2013bundled,zhang2016robust}. In this work, we propose deep structured models to produce content-aware flow fields for image warping.  

\begin{figure*}[t]
	\begin{center}
		\includegraphics[width=\textwidth]{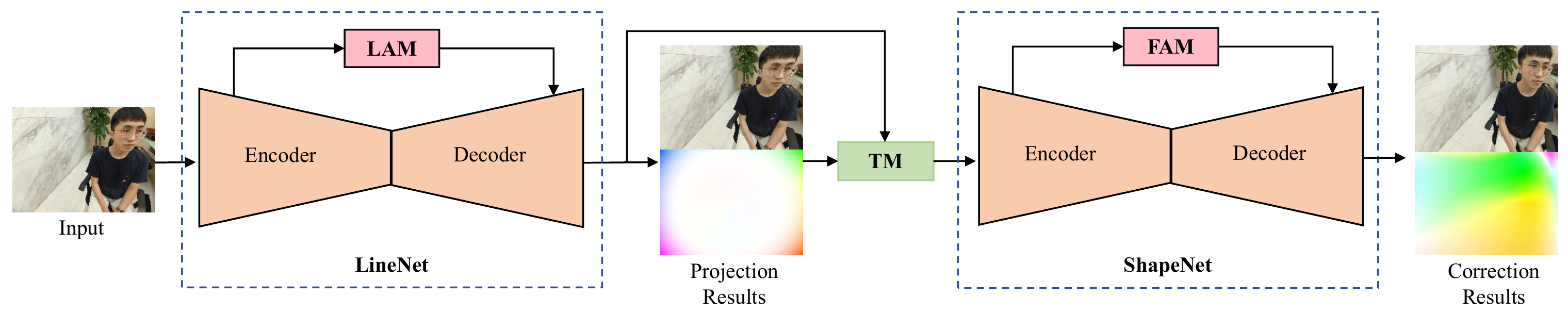}
	\end{center}
	\vspace{-3mm}
	\caption{An overview of our network architecture.}
	\vspace{-3mm}
	\label{fig:structure}
\end{figure*}

\subsection{Face Undistortion}
Several methods are proposed to correct distorted faces.\cite{Beeler2010,Bryan2012,Artizzu2014}. Fried~\emph{et al.} proposed to fit a 3D face model to modify the relative pose between the camera and the subject by manipulating the camera focal length, yielding photos with different perspectives~\cite{fried2016perspective}. A more related work is proposed by Shih~\emph{et al.}~\cite{shih2019distortion}, which corrects wide-angle portraits by content-aware image warping. However, it requires the camera parameters and results in distortion either on the background or on the faces.
%In addition, the transitions are not smooth, resulting in distortion either on the background or on the faces. 
In contrast, our method is calibration-free and achieves smooth transitions with a deep neural network. 

%methods：
\section{Methodology}
The proposed network contains two sub-networks, line correction network (LineNet) and portrait correction network (ShapeNet), Fig.~\ref{fig:structure} shows the overall architecture and the pipleline of the wide-angle image correction. As can be seen, the first LineNet generates a perspective projection flow from the given distortion image to project the image as flattened. Then the ShapeNet predicts the face correction flow from the flattened image. In order to make the two networks work on different deformations, we design self attention modules LAM and FAM to work on the two networks respectively. These two sub-networks are bridged by a transition module TM. Finally, the projection image is transformed to correction image with correction flows.

\subsection{Line Correction Network}
%The main task of Line Correction Network is to restore a flattened image from the perspective distortion input. Shih \emph{et al.}~\cite{shih2019distortion} achieves this by adopting distortion parameters. However, in our method, we have no such parameters, where different modules have different distortions. We propose a LineNet to learn such correction blindly, purely based on our training data. Since the purpose of the projection is to keep straight lines, we add an additional attention module to predict straight lines in the network. 

\begin{figure*}[t]
	\begin{center}
		\includegraphics[width=0.98\textwidth]{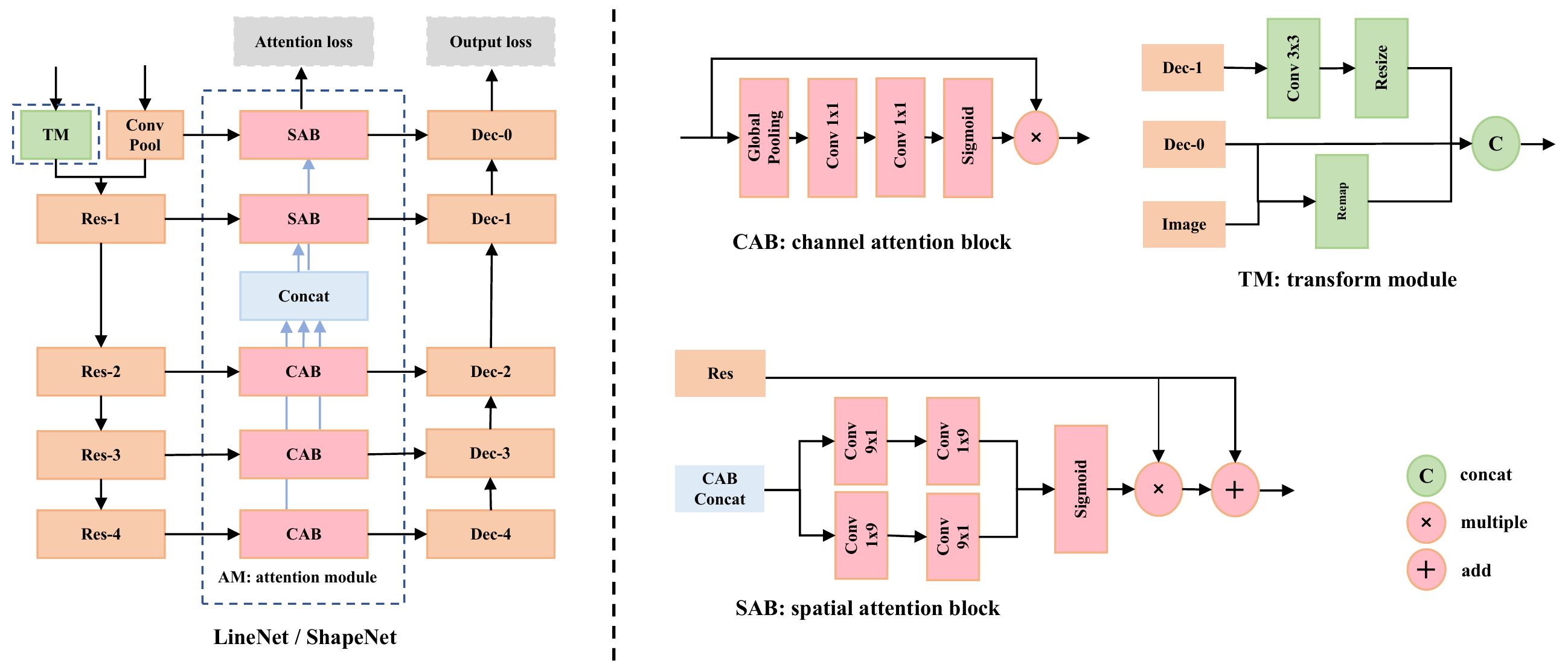}
	\end{center}
	\vspace{-3mm}
	\caption{The overall structure of our model. \textbf{ShapeNet} and \textbf{LineNet} share the similar structure, with \textbf{TM} avaliable only for \textbf{ShapeNet}.}
	\vspace{-3mm}
	\label{fig:submodels}
\end{figure*}

\textbf{LineNet} As shown in Fig.~\ref{fig:submodels}, a standard encoder decoder network is used to predict the corresponding deformed flow from a single image. It consists of two phases, namely, down-scale feature maps and up-scale feature maps. Given an original image ${I}$, we adopt \textit{ResNeXt}~\cite{xie2017aggregated} network as the backbone to extract feature maps. Then, the feature pyramid is input into up-scale decoders to generate the final flows. In each decoder, the feature maps are inferred by two $3 \times 3$ convolutional layers, an add operator with the previous encoded feature, and two additional $3 \times 3$ convolutional layers. Then, a deconvolution operator is adopted to up-scale the decoder feature map. After that, the projection image is obtained through the flow-based warping.

\textbf{LAM} Since the main purpose of the LineNet is to straighten the bended lines as shown in Fig.~\ref{fig:structure}, we design a line attention module (LAM) to learn various lines. Different from other multi-branch methods, we use the intermediate results of encoder and decoder for prediction. LAM consists of two main blocks, channel attention block (CAB) and spatial attention block (SAB). As shown in Fig.~\ref{fig:submodels}, CAB is applied on high level features to generate the attention in channel perspective. It contains a global pooling operation and two convolution layers to generate a $C\times 1$ attention map from a given feature map $C\times H\times W$. All the high-level CAB outputs are concatenated into the following SAB. SAB combines the original encoder features and CAB features to output the final spatial attention feature maps. 

In order to locate the lines in the image, we apply the two lowest level SAB outputs as additional line supervisions. Therefore, we use Sobel operator to extract the corresponding edge from the original image as the ground truth, and then calculate the loss between the edge map and the SAB output. LAM has two advantages. On one hand, it uses the information of high-level and low-level to strengthen the global and local features of the image. On the other hand, the edge constraint makes the encoder and decoder feature pay more attention to the edges. Without additional calculations, the model of line correction becomes feasible. 

\subsection{Portrait Correction Network}
\textbf{ShapeNet} After obtaining the results of line correction, we need to further produce the flow of face correction. We use the same network architecture as LineNet. Portrait Correction Network also outputs a flow map, but it aims to correct the face areas while leaving the background unchanged.

\textbf{FAM} Different from the LAM of the first network, the main purpose of ShapeNet is to rectify the face area. Because the face is only a small part of the image, when the face is deformed, the boundary between the face and the background would inevitably be distorted. In order to describe the transition region more accurately, the most direct method is to segment the human image to obtain the accurate head boundary. However, depending on the energy transfer of the attention module, accurate segmentation is not the most necessary. Instead, we generate a heatmap of human face based on the results of face detection to adaptively learn the changes of face and transition region, as shown in Fig.~\ref{fig:dataset} (d). In the same way, face heatmap is used as the supervisions of face areas. 

\textbf{TM} Considering that while ShapeNet is performing face correction, the perspective projection transformation from the LineNet should be maintained, so in order to make it easier for LineNet and ShapeNet to keep the consistency of the non-portrait area, we proposed a transition module (TM) to transfer the distortion from the LineNet to ShapeNet. 
%Although the two networks work on different regions of the image, they could lead to inconsistent deformation inevitably. %Based on this, we introduce the line morphing featmap into the second network.

The TM has three parts, as shown in Fig.~\ref{fig:submodels}. The first part is the decoder feature maps of the penultimate layer, following with convolution and up sampling. The second part is the final flow map, and the third part is the projection image. Three feature maps are concatenated together as inputs to the ShapeNet. This module contains image features, location features, as well as hidden semantic information.

\subsection{Training and Inference} 
\textbf{Loss Function.} Our model is learned in an end-to-end manner. For each sub-network, we adopt L2 loss between the generated and ground truth of flows and image respectively. Besides, we apply a boundary preservation L2 loss to enhance the edge accuracy. Sobel operator is adopted to generate the edge of ground truth and predication.
%Then the same L2 loss is used to supervise the reconstruction of boundaries. The similar boundary supervision is used and proved in several previous works. Different from them,
Different from boundary supervision used in several previous works, Sobel operator can also smooth flows to avoid aliasing in the undistorted images. The formulas are as follows:
\setlength\abovedisplayskip{3pt plus 3pt minus 6pt}
\setlength\belowdisplayskip{3pt plus 3pt minus 6pt}
\begin{equation}
L_{line} = ||F_{flow} - I_{flow}||_{2, s2} + ||F_{proj} - I_{proj}||_{2, s2}
\end{equation}
\begin{equation}
L_{shape} = ||F_{flow} - I_{flow}|_{2, s2} + ||F_{out} - I_{out}||_{2, s2}
\end{equation}
where $L_{line}$, $L_{shape}$ are the losses of the output of LineNet and ShapeNet. Two kinds of losses, L2 and $sobel_L2$ are used on both flow and images. Similar loss is also applied to the attention modules.
\begin{equation}
L_{LAM} = ||F_{lam} - I_{edge}||_2
\end{equation}
\begin{equation}
L_{FAM} = ||F_{fam} - I_{face}||_2
\end{equation}
$L_{LAM}$ is the difference between LAM output and the provided edge ground truth, while $L_{FAM}$ is the difference between FAM output and the labeled face heatmap. The total loss function is the weighted sum of the above losses. 
\begin{equation}
L = \lambda_1 L_{LAM} + \lambda_2 L_{FAM} + \lambda_3 L_{line} + \lambda_4 L_{shape}
\end{equation}
The $\lambda_{1,2,3,4}$ is used to balance the importance among the reconstruction and attention losses. we set them to 5, 5, 1, 1 respectively in all experiments.

\textbf{Inference.} In the inference stage, the generated two flows %generated in the structured networks 
can be combined into one flow to describe the offset directly from the original image to the final corrected. Given an image, it is first reduced to $256\times 384$ to obtain two flow maps of $256\times 384$. After fusion, the fused flow is resized to the original size to generate the correction image.

\section{Data Preparation}
There is no dataset of paired portraits. We therefore create a novel training dataset by ourselves which contains distorted and undistorted image pairs with various camera modules, scenes and identities. We used $5$ different ultra wide-angle camera of smartphones, and photographed over $10$ people in several scenes. The number of people in each photo ranges from 1 to 6. Overall, over $5,000$ images were collected. Given a distorted portrait image, it is non-trivial to obtain its corresponding distortionless image. We propose to correct wide-angle portrait into a distortion-free image manually. To achieve this, we propose to first run an improved Shih's~\cite{shih2019distortion} algorithm iteratively and then further improve the results by our designed manual tool.

\begin{figure*}[htbp]
	\begin{center}
		\includegraphics[width=\textwidth, height=95pt]{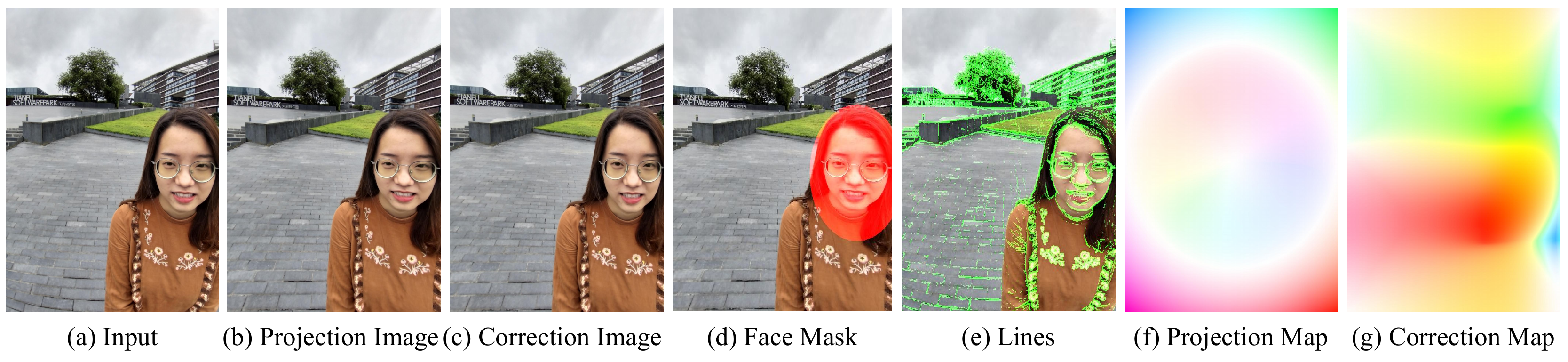}
	\end{center}
	\vspace{-3mm}
	\caption{Training example. (a)input image. (b)projection image with straight lines. (c)corrected image by enhanced Shih's method~\cite{shih2019distortion} and our manual tool. (d)face mask created by face detection used in FAM. (e)lines created by line detection used in LAM. (f)projection flow from (a) to (b), which is the guidance of LineNet. (g)correction flow from (b) to (c), which is the guidance of ShapeNet.}
	\vspace{-2mm}
	\label{fig:dataset}
\end{figure*}

\begin{figure}[htbp]
	\centering
	\includegraphics[width=0.95\linewidth]{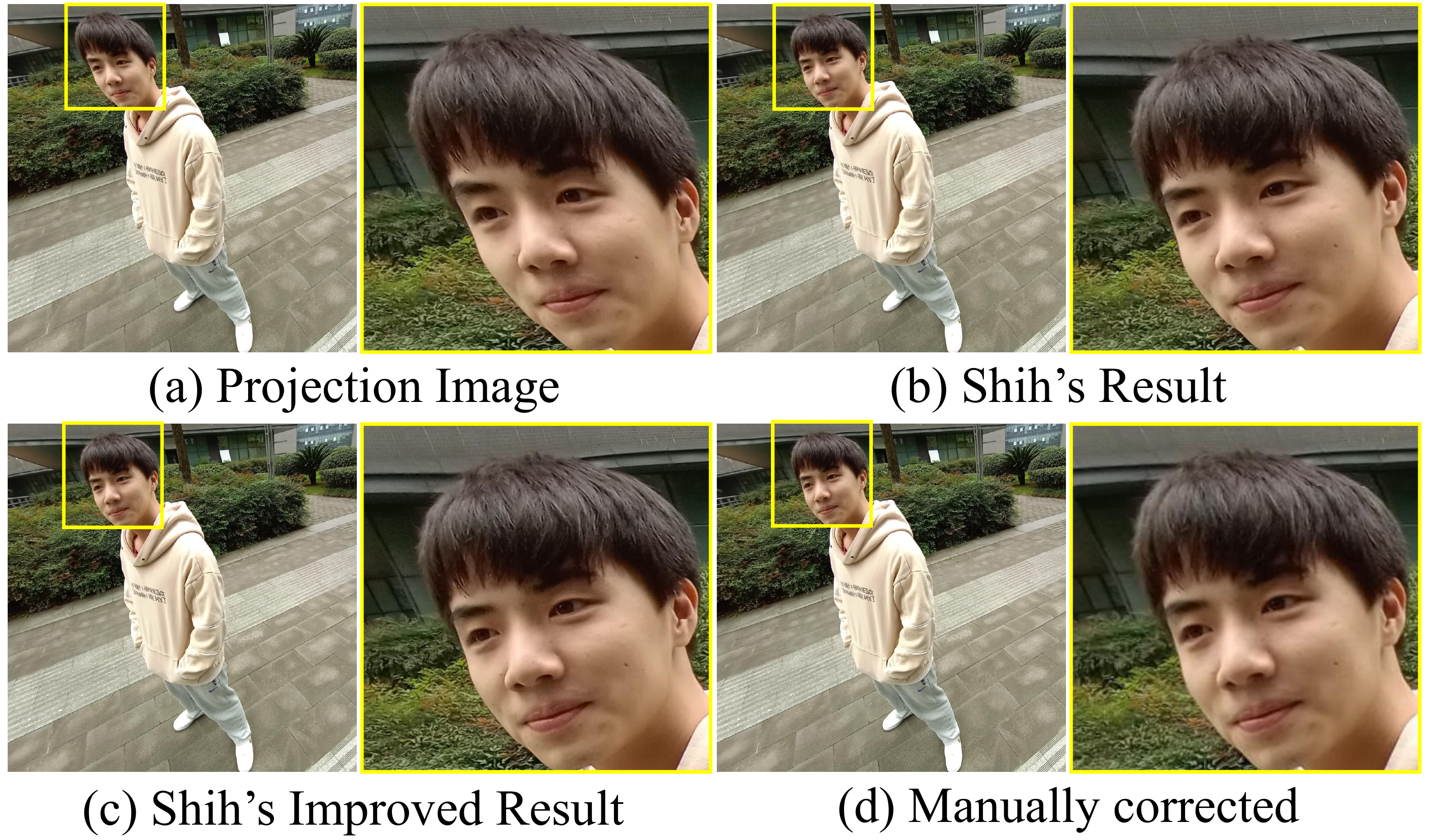}
	\caption{Example of our dataset creation. We built a tool that can adjust the warping of meshes for ground-truth generation. We use the results from Shih~\emph{et al.}~\cite{shih2019distortion} as input, and correct their problematic regions manually. (a) input image and the cropped region. (b) result by Shih~\emph{et al.}~\cite{shih2019distortion}. Please notice the bending of lines and unnatural shape of the face. (c)Shih~\emph{et al.}~\cite{shih2019distortion}'s improved result. (d) our manually corrected results of (c). }
	\vspace{-3mm}
	\label{fig:tool}
\end{figure}

We notice that the edges near the faces are often distorted in~\cite{shih2019distortion}. We improve the results of ~\cite{shih2019distortion} by adding explicit line constraints~\cite{lin2016seamless}. Fig~\ref{fig:tool} (a) is the input image. Fig~\ref{fig:tool} (b) is Shih's result~\cite{shih2019distortion}, and Fig~\ref{fig:tool} (c) is our improved result. Notably, this improvement can alleviate the problem to some extent, but cannot be perfect. Specifically, we run our improved method iteratively in the following steps: 1) run the algorithm and obtain the initial results; 2) for results that look unnatural, we re-run the algorithm by adjusting the hyper-parameters. 3) Repeat step 2 until the correction results converge. In the experiment, about half of the images can be satisfactory after the first optimization, and the remaining half need $5$ more iterations to complete. Finally, the image which cannot get good results is discarded.

Now, we obtain an initial dataset. To further improve the quality, we correct some unsatisfactory parts by our designed manual tool. Our manual tool is mesh-based, with as-rigid-as-possible quads constraints~\cite{liu2009content} and line-preserving constraints~\cite{lin2016seamless}. Users can drag the image content by a mouse to drive the mesh deformation interactively. As shown in Fig.~\ref{fig:tool} (d), we further correct the result of (c) for improvements.  Fig.~\ref{fig:dataset} shows an example. The flow motions, Fig.~\ref{fig:dataset} (f) and (g), can be obtained during our manual correction, which are used as the guidance to the network LineNet and ShapeNet, accordingly. Face mask and lines, Fig.~\ref{fig:dataset} (d) and (e), are used for LAM and FAM, accordingly. 
% We plan to release the dataset.

\section{Metrics}
In this section, we introduce two novel evaluation metrics: Line Straightness Metric (LineAcc) and Shape Congruence Metric (ShapeAcc). As far as we know, there is no suitable quantitative metric in the field of distortion correction. The accuracy of quantitative calculation needs a corrected image as a reference, where our manually corrected images are used. 

\begin{figure}[t]
	\centering
	\includegraphics[width=0.98\linewidth]{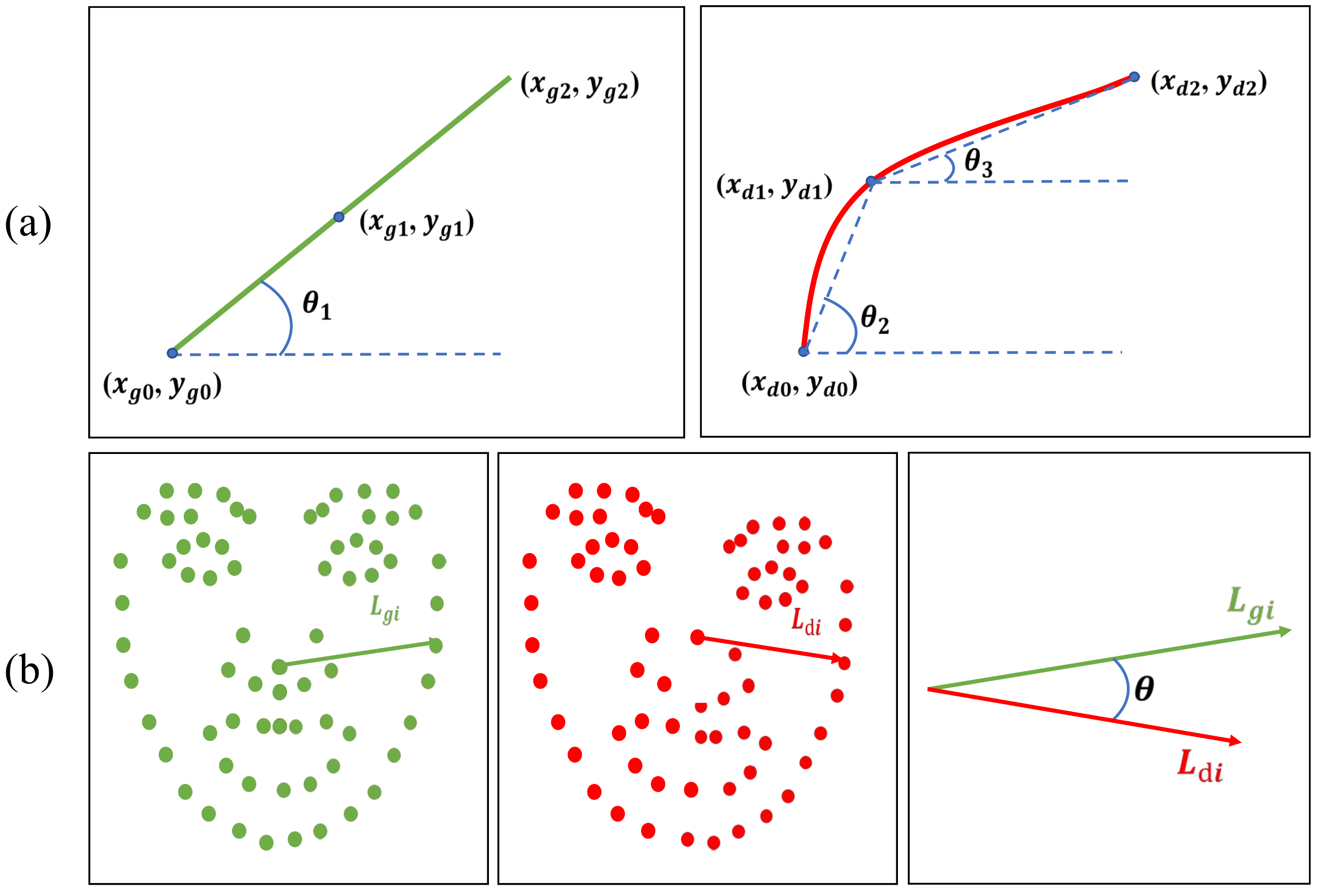}
	\caption{(a) Line Straightness metric (LineAcc) and (b) Shape Congruence Metric (ShapeAcc). }
	\label{fig:metric_calculation}
\end{figure}

\begin{figure}[t]
	\centering
	\includegraphics[width=1.0\linewidth]{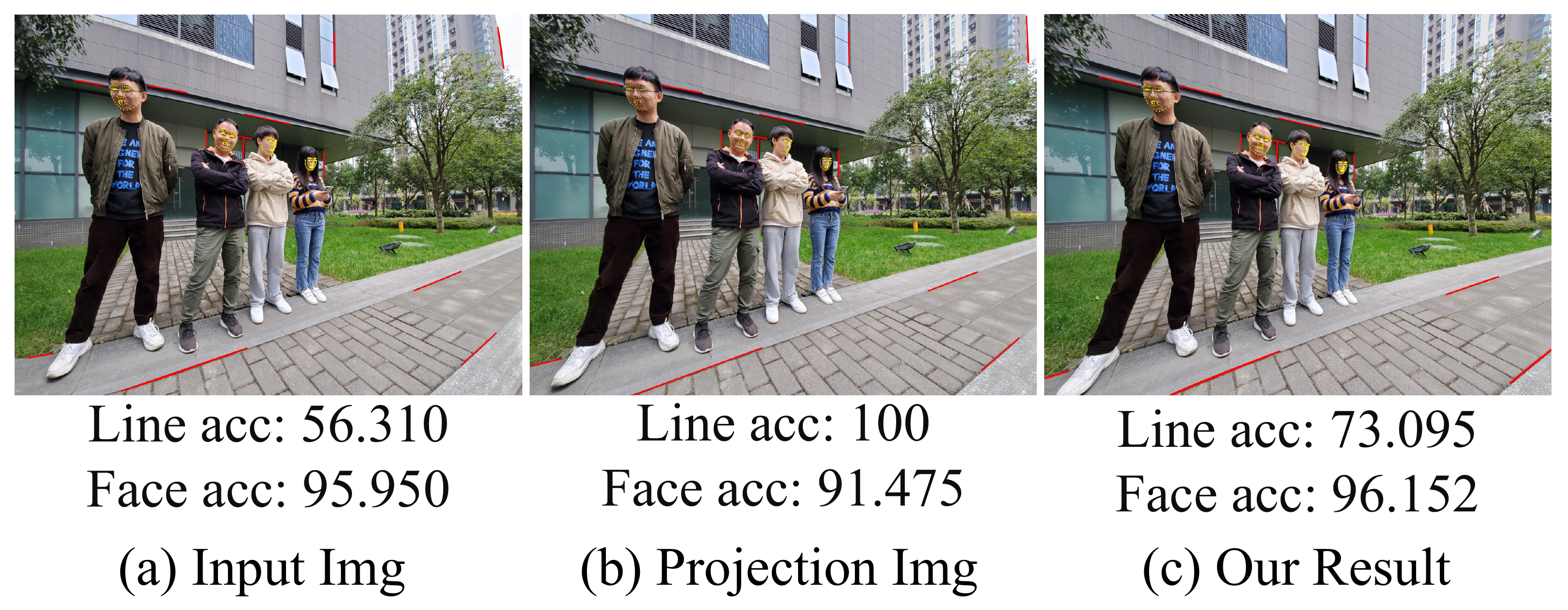}
	\caption{(a) input image with LineAcc and ShapeAcc. (b) Perspective undistorted image, where LineAcc increases and ShapeAcc decreases. (c) Our result with two scores balanced, both of which are increased compared with (a).}
	\label{fig:twometrics_comp}
\end{figure}

\noindent\textbf{Line Straightness Metric}: 
The salient lines should keep straight after correction. We mark salient lines in the test dataset and then calculate the curvature variation of the marked lines. For each line $L$, we uniformly sample $n$ points $p_0$, $p_1$, ..., $p_n$. Then, the slope variation of the line can be calculated as: 
\begin{equation}
S_{g} = \frac{y_{g_0} - y_{g_n}}{x_{g_0} - x_{g_n}}
\end{equation}
\begin{equation}
LS = 1 - (\frac{1}{n} \sum_{i=0,..,n-1} [\frac{y_{d_i} - y_{d_{i-1}}}{x_{d_{i}} - x_{d_{i-1}}} - S_{g}])
\end{equation}
%\begin{equation}
%S_{d} &= \frac{1}{n} \sum_{i=0,..,n-1} [\frac{y_{d_i} - %y_{d_{i-1}}}{x_{d_{i}} - x_{d_{i-1}}} - S_{g}]
%\end{equation}
%\begin{equation}
%LS &= S_{g} - S_{d},
%\label{equ:coorlation}
%\end{equation}

\noindent where $P_{g_{i}} = [x_{g_{i}}, y_{g_{i}}]$, $P_{d_{i}} = [x_{d_{i}}, y_{d_{i}}]$ depicts the location of corresponding point in reference and distortion images. $LS$ is the similarity between slope of these two lines. The line is more curved, the value of LineAcc is smaller. Fig.~\ref{fig:metric_calculation} (a) shows an illustration. We mark the salient lines near the image boundary and around the portraits.

\noindent\textbf{Shape Congruence Metric}: 
Given a face, we label the landmarks on the result and the reference image, and then calculate the similarity between the two groups of landmarks. Fig.~\ref{fig:metric_calculation} (b) shows an example, where the vectors are produced from the nose landmark to other landmarks. 
\begin{equation}
FC = \frac{1}{n} \sum_{i=0,..,n-1} [cos({L_{g_i}, L_{d_i}})]
\end{equation}
\begin{equation}
cos({L_{g_i}, L_{d_i}}) = \|L_{g_i}\|\|L_{d_i}\| cos\theta
\end{equation}
\noindent where $L_{g}$ and $L_{d}$ depict the corresponding landmarks in the reference image and the result image. Fig.~\ref{fig:twometrics_comp} shows an example of the two metrics. Fig.~\ref{fig:twometrics_comp} (a) is the input with original LineAcc and ShapeAcc. Fig.~\ref{fig:twometrics_comp} (b) is the line corrected by perspective undistortion, where the LineAcc increases, and the ShapeAcc decreases. Fig.~\ref{fig:twometrics_comp} (c) is our result. We achieve a balanced scores regarding the two metrics. Note that, the dimensions of the two metrics are not the same.  

\section{Experiments}
In this section, we first analyze the impact of each module by the ablation study, and then based on the best experimental module configuration, we conduct another two data ablation experiments on the training set of different phone modules to verify the generalization of the network. Finally, we make a quantitative and qualitative comparison with the related works, and also compare with some smartphones with wide-angle correction. %~\cite{shih2019distortion} and Stereographic Projection, 

\noindent\textbf{Implementation details} All experiments are carried out on our training set. We use the standard data augmentation procedure to expand the diversity of training samples, including horizontal flip, random crop and scaling. Besides, we sample uniformly according to the number of people in the image, and increase the proportion of people in corner to solve the problem of imbalance between the portraits at the center and corner. 
%Given training images, we randomly crop the $256 \times 384$ patches as inputs. 
We use the ADAM optimizer \cite{Adam} with an initial learning rate of 5e-3 , which is divided by $10$ at the 150-th epoch, and the total training epoch is $200$.
%The hardware configuration are $4$ NVIDIA RTX 2080Ti GPUs and $4$ Intel Xeon Gold 6130 2.10 GHz CPUs. 

\noindent\textbf{Test set} We construct a test set that contains both face landmarks and lines near the faces and at the corners of the image to appreciably evaluate the wide-angle portrait correction. It contains a total of 129 original wide-angle images of 5 different phone modules and the corresponding calibrated images according to camera parameters. Experiments are evaluated on our test set and Shih's~\cite{shih2019distortion} test set.

\subsection{Ablation Studies}
In this subsection, we step-wise decompose our approach to reveal the effect of each component. The basic LineNet is built on the straightforward encoder-decoder structure, which takes the original distorted image as input, and predicts the perspective projection flow map. Based on this, we verify the function of LAM, ShapeNet, TM and FAM module respectively. The LineAcc and ShapeAcc are adopted for the evaluation. LineAcc is evaluated on results of both LineNet and ShapeNet 
%and reported as Projection LineAcc and Correction LineAcc 
in Table.~\ref{table:abs}. As seen, all the proposed modules contribute to the performance. 

\begin{table*}[t]
	\centering
	\caption{Ablation study of the proposed network. "LineNet", ”LAM”, ”ShapeNet”, "TM" and "FAM" refer to the basic LineNet, Line Attention Module, ShapeNet, Transtion Module, Face Attention Module, respectively. "Lmk Loss" refers to the adaption of landmark loss onto ShapeNet. "Proj LineAcc", "Corr LineAcc", and "ShapeAcc" refer the Projection and Correction LineAcc Metric of two networks and the final Shape Correction Accuracy. In addition, results of three methods are adopted for comparisons. They are Input Image, projection result with calibrated params, and Shih's~\cite{shih2019distortion} result. The best is marked in {\color{red}red} and the second best is marked in {\color{blue}blue}.}
	\vspace{3mm}
	\setlength{\tabcolsep}{8 pt}
	\begin{tabular}{c|ccccc|ccc}
		\hline
		No. & LAM & ShapeNet & TM & FAM & Lmk Loss & Proj LineAcc & Corr LineAcc & ShapeAcc\\
		\hline
		1) LineNet & & & & & & 66.745 & $\setminus$ & 97.380 \\
		2) & $\checkmark$& & & & & 66.856 & $\setminus$ & 97.391 \\
		3) & & $\checkmark$& & & & 66.707 & 66.439 & 97.458 \\
		4) & $\checkmark$&$\checkmark$ & & & & 66.873 & 66.472 & 97.472 \\
		5) & $\checkmark$& $\checkmark$& &$\checkmark$ & & 66.938 & 66.484 & 97.479 \\
		6) & $\checkmark$& $\checkmark$& & & $\checkmark$& 66.985 & 66.541 & 97.473 \\
		7) & $\checkmark$& $\checkmark$& $\checkmark$& $\checkmark$& & {\color{blue}67.069} & {\color{blue}66.575} & {\color{blue}97.485} \\
		8) & $\checkmark$& $\checkmark$& $\checkmark$& $\checkmark$& $\checkmark$& {\color{red}67.135} & {\color{red}66.784} & {\color{red}97.490} \\
		\hline
		\hline
		9) Input & & & & & & 66.064 & 66.064 & 97.455 \\
		\hline
		10) Proj Img & & & & & & $\setminus$ & $\setminus$& 96.876 \\
		\hline
		11) Shih~\cite{shih2019distortion} & & & & & & 66.143 & 66.143 & 97.253\\
		\hline
	\end{tabular}
	\label{table:abs}
\end{table*}

\begin{table*}
\footnotesize 
\caption{Quantitative comparisons of ours and Shih~\cite{shih2019distortion} on different test sets. The first three rows in the table indicate the models training without note data, without vivo data and with the total training set, respectively. And correspondingly test on note, vivo, ours whole test and google test. 
The best two scores are shown in {\color{red}red} and {\color{blue}blue}.}
\vspace{-0.5em}
\begin{center}
\setlength{\tabcolsep}{10 pt}
\begin{tabular}{c | c c | c c | c c | c c }
\hline
\multirow{2}{*}{No.} &
\multicolumn{2}{c}{note testset} & 
\multicolumn{2}{c}{vivo testset} &
\multicolumn{2}{c}{all testset} &
\multicolumn{2}{c}{google} 
\\
\cline{2-9}
& LineAcc & ShapeAcc & LineAcc & ShapeAcc & LineAcc & ShapeAcc & LineAcc & ShapeAcc \\
\hline
\hline
1) ours wot note  & 67.605&  97.061& \color{blue}64.997 & 98.341 & \color{blue}66.464 & 97.464 & $\setminus$ & $\setminus$  \\
2) ours wot vivo  & \color{blue}68.299 & 97.109 & 63.418 & \color{blue}98.361 & 66.324 & \color{blue}97.483 & $\setminus$ & $\setminus$ \\
3) ours with all & \color{red}68.683 & \color{blue}97.115 & \color{red}65.148 & \color{red}98.363 & \color{red}66.784 & \color{red}97.490 & \color{red}64.650 & \color{red}97.499 \\
\hline
\hline
4) google(Shih~\cite{shih2019distortion}) & 66.886 & \color{red}97.267 & 63.087 & 98.238 & 66.143 & 97.253 & \color{blue}61.551 & \color{blue}97.464 \\
%stereo &  & $\setminus$ &  & $\setminus$ & & $\setminus$ &  & $\setminus$ \\
\hline
\end{tabular}
\end{center}
%\vspace{-1mm}
\label{tab:Quantitative}
\end{table*}

\noindent\textbf{LineNet} The basic LineNet improves the Projection LineAcc from $66.064$ to $66.745$, which is obviously beyond Shih's~\cite{shih2019distortion} approach. 

\begin{figure}[t]
    \centering
    \includegraphics[width=1.0\linewidth]{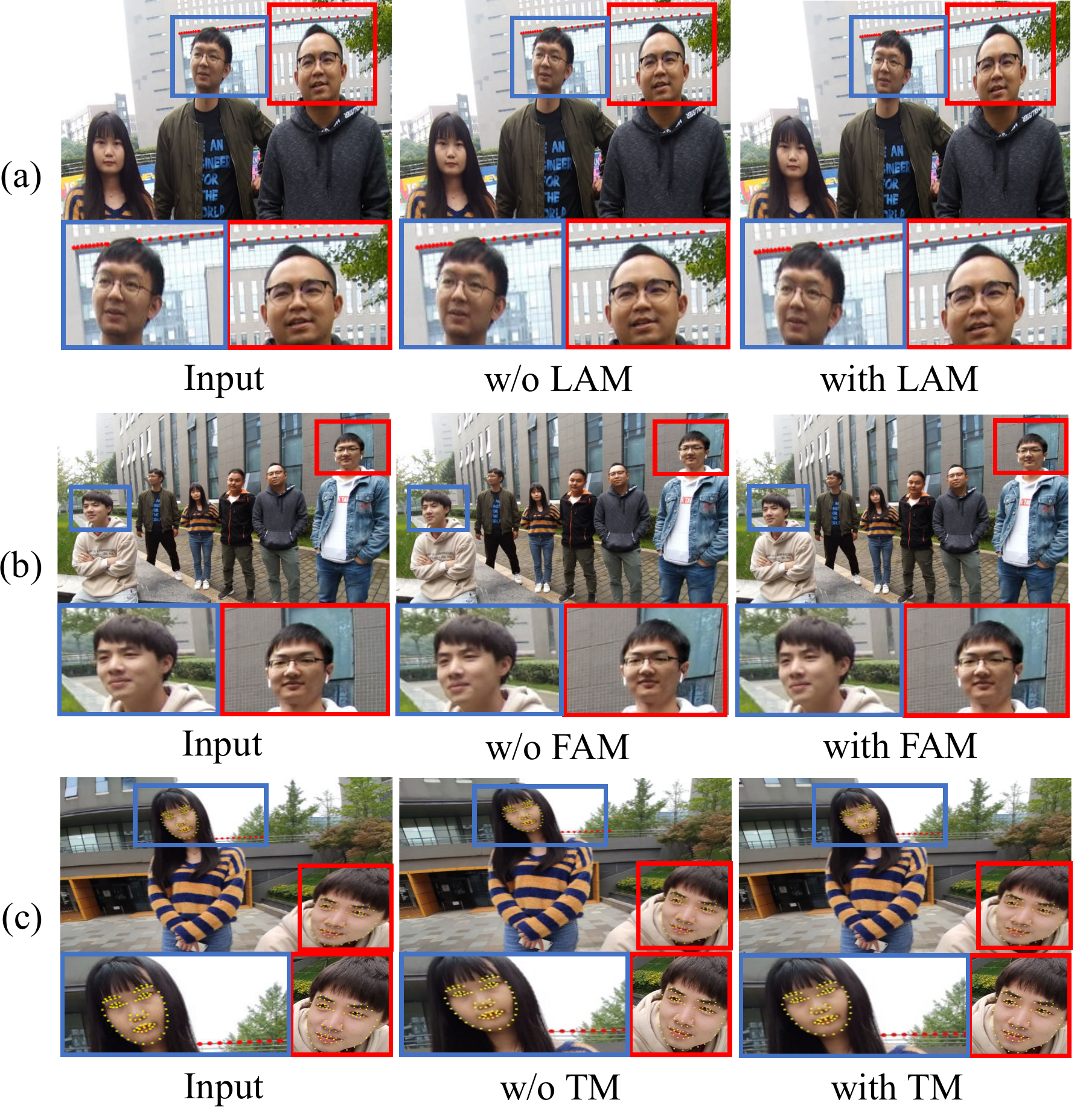}
    \caption{Visualization of ablation study. (a), (b), (c) represent the different performance of the network with or without LAM, FAM and TM, and prove their effects respectively.}
    \label{fig:FAM_LAM_TM}
\end{figure}

\noindent\textbf{LAM} Compare 1) and 2) in Table~\ref{table:abs}, after integrating the Line Attention Module with the basic LineNet, the LineAcc is improved from $66.745$ to $66.856$, because the Line Attention Module facilitates the learning of line-awareness features. As shown in Fig.~\ref{fig:FAM_LAM_TM} (a), the straight line in the corner is obviously straighter, with the constraint of LAM.

\noindent\textbf{ShapeNet} Next, we integrate the Basic ShapeNet with LineNet. The evaluation result can be seen in 3) of Table.~\ref{table:abs}. ShapeNet affects the line accuracy of LineNet to a certain extent, but obviously significantly improves the accuracy of face correction. Furthermore, LAM shows strong robustness and improves the final line calibration accuracy in the end-to-end network by comparing 3) and 4).

\noindent\textbf{FAM} Furthermore, FAM is applied onto ShapeNet. As shown in 4) and 5) of Table.~\ref{table:abs}, compared to the standard structure, FAM improves the ShapeAcc Metric from $97.472$ to $97.479$, which is due to the improved confidence of the face areas. The same conclusion can be found in Fig.~\ref{fig:FAM_LAM_TM} (b). 
%Because of the constraint of LAM, the deformatifon of face area becomes more natural and harmonious.

\noindent\textbf{TM} The purpose of TM is to further balance the deformation of salient straight lines and faces. As depicted in 7) in Table.~\ref{table:abs}, the correction LineAcc and ShapeAcc are improved from $66.484$ to $66.575$, and from $97.479$ to $97.485$, respectively. As shown in Fig.~\ref{fig:FAM_LAM_TM} (c), the transition area between head and background is more natural while TM transfers the line projection features to ShapeNet. 

\noindent\textbf{Lmk loss} Finally, we verify the effectiveness of the landmark loss. Experimental results show that  landmark loss does not bring a particularly large improvement in the accuracy of face correction, because FAM has improved the accuracy of face correction with a relatively large margin. But on the other hand, it improves the accuracy of line correction. This is mainly due to more accurate face edge constraints, which smooths the transition at face boundary regions. Based on the superposition of the above modules, the accuracy of the proposed model is much higher than that of the Shih's~\cite{shih2019distortion}, in terms of both LineAcc and ShapeAcc.

\begin{figure*}[htbp]
	\begin{center}
		\includegraphics[width=1.0\textwidth]{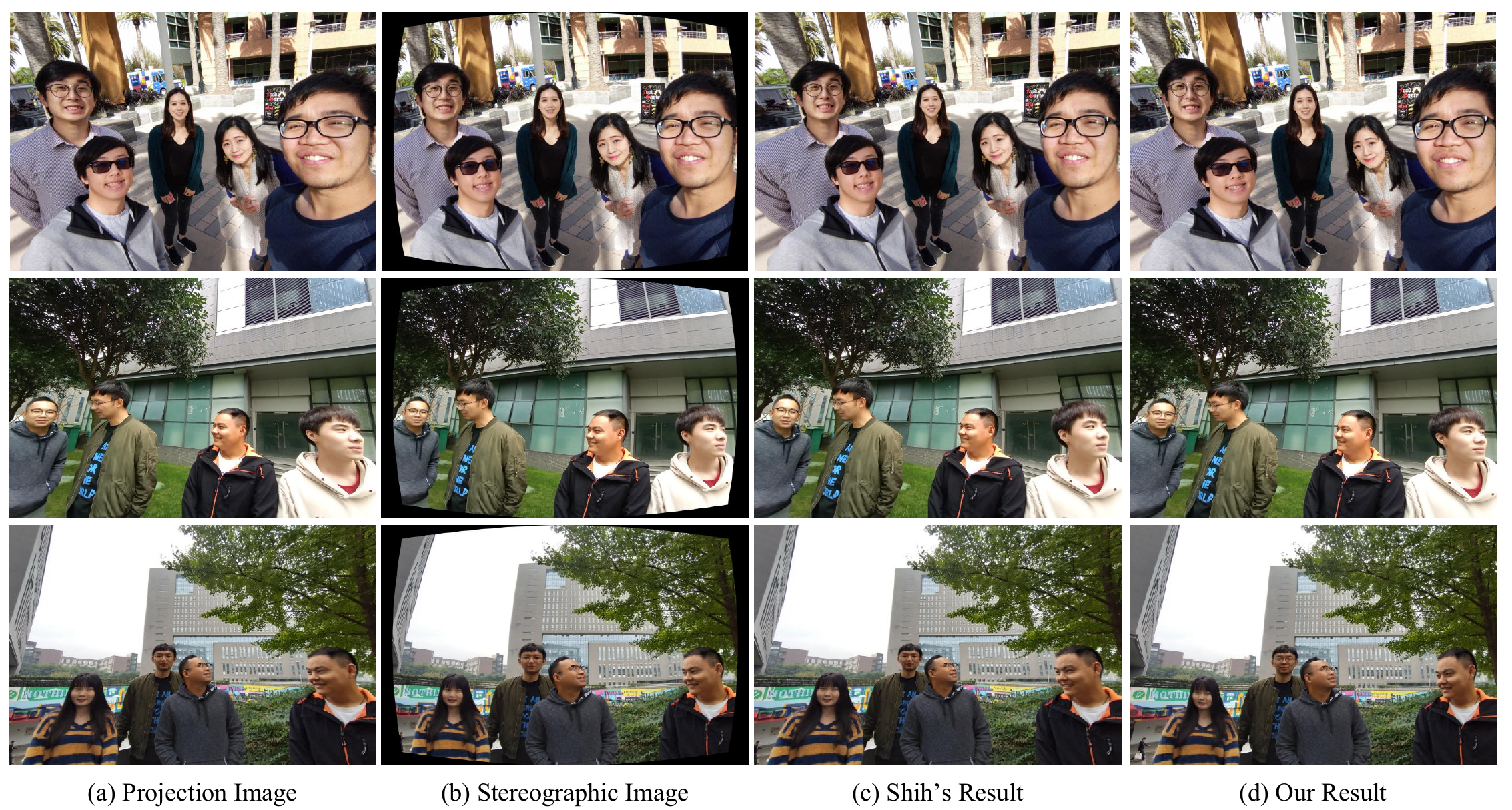}
	\end{center}
	%\vspace{-2mm}
	\caption{Qualitative evaluation of undistortion methods. Notice the coordination of face area and line curvatures in the transition area.}
	%\vspace{-1mm}
	\label{fig:compare_with_other_methods}
\end{figure*}

\begin{figure*}[htbp]
	\begin{center}
		\includegraphics[width=1.0\textwidth]{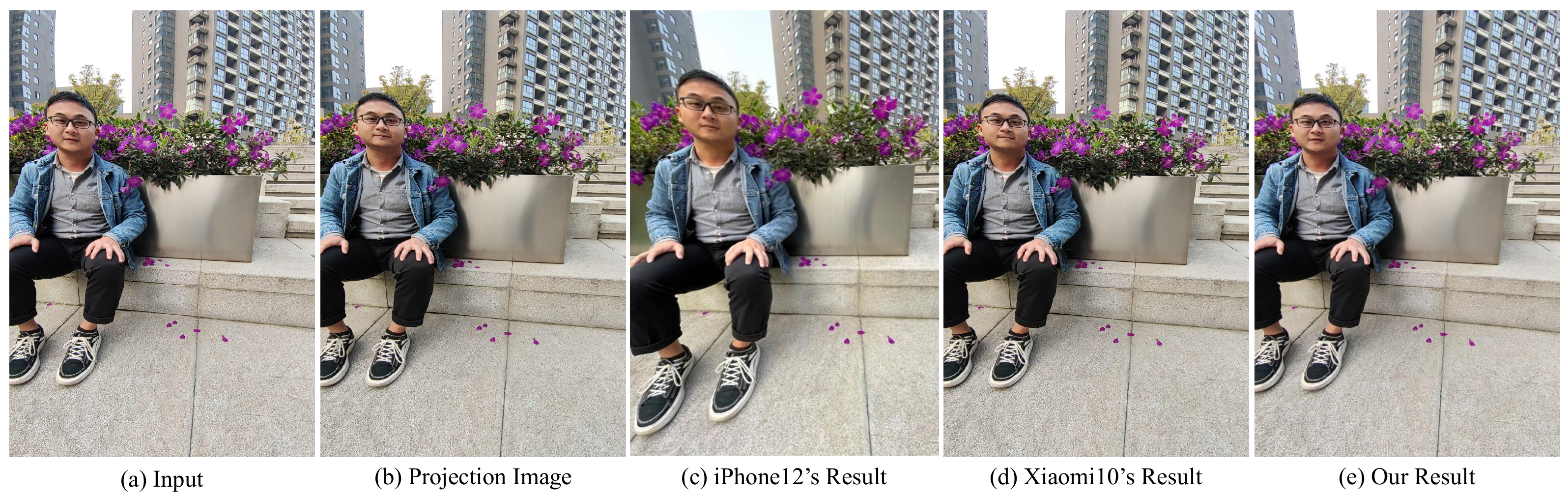}
	\end{center}
	%\vspace{-2mm}
	\caption{Qualitative comparison between our method and some phones with wide-angle portrait correction.}
	%\vspace{-1mm}
	\label{fig:phones}
\end{figure*}

\noindent\textbf{Generalization verification} Based on the model 8) in Table.~\ref{table:abs}, we select two smartphone modules from the training and testing set respectively for cross testing, and the full test set and test set of ~\cite{shih2019distortion} are also used for the final evaluation to verify the robustness and generalization of the network. As shown in Table.~\ref{tab:Quantitative}, comparing 1) 3) and 2) 3), adding data of different models can improve the robustness. At the same time, except for the ShapeAcc on note testset, our methods perform better than Shih~\cite{shih2019distortion} on almost all 4 test sets, which indicates the good performance of generalization.
%2) achieved the highest ShapeAcc on vivo testset in the absence of vivo data, and our methods perform better than Shih~\cite{shih2019distortion} on both 4 test sets, which indicates the good performance of generalization.

\subsection{Comparison with Other Methods}
%We compare the proposed network with stereographic projection method and Shih~\cite{shih2019distortion}, since stereographic projection can project portraits with the minimal distortion, and Shih~\cite{shih2019distortion} rectify the background and portraits simultaneously.
% The performance of line and shape accuracy are depicted in Table.~\ref{tab:Quantitative}. We can see that our method improves visibly in terms of LineAcc and ShapeAcc on both Shih's testset and our proposed testset. When using different training sets, it is still significantly better than the opponent. 

Fig.~\ref{fig:compare_with_other_methods} shows the visual comparisons with Shih's results~\cite{shih2019distortion}. The projection image can correct lines but cause serious distortion on face regions, while the stereographic projection can maintain the faces but suffer from structure bending. Shih's results~\cite{shih2019distortion} can seek a balance between the faces and the background. However, some structures are still bended at the background, and some faces still suffer from a bit distortion, unbalanced with the body. In contrast, our results are more natural in the correction of the head, and the transition area between the portrait and the background is smoother. The line in the background is also closer to the result of perspective projection, and the faces look more natural. More notably, in the second row of Fig.~\ref{fig:compare_with_other_methods}, our results can correct the rightmost face while keeping the architectural lines above it still straight, while the lines
above the face in Shih's~\cite{shih2019distortion} are obviously deformed. 
Metrics in Table.~\ref{table:abs} and Table.~\ref{tab:Quantitative} also confirm the conclusion, since the accuracy in terms of both LineAcc and ShapeAcc has been greatly improved from Shih's~\cite{shih2019distortion}.

\subsection{Comparison with Other Phones} Furthermore, we compare the results with some smartphones with wide-angle portrait correction. Two flagship phones of Xiaomi and iPhone are applied as comparisons. As shown in Fig.~\ref{fig:phones}, there is serious stretching of portraits and some bending of background in the result of iPhone 12. The result of Xiaomi 10 is close to the result of perspective projection, and there is still slight deformation of the face. Our results are significantly better than others, as the face is undistorted while correcting the background lines. 
%We can compress the model to about 400M FLOPs with almost no loss of accuracy, and achieve the real-time image rectification on mobile phone.

% \subsection{Visualization of Corner Cases}
More comparisons are shown in our project page, including corrections for photos from the Internet, as well as some failure cases.

% 需要补充图像，和优化文字。体现出来1、背景和人像的精度都更高；2、flops是多少，运行速度可以达到实时。

\section{Conclusion}
% 需要补充结论 和 failure cases
%\vspace{-2mm}
This paper proposes a deep structured network to automatically correct distorted portraits in wide-angle photos, which applies perspective projection to background and stereographic projection to portraits, and achieves a smooth transition between them. It does not rely on any prior calibrated parameters, thus being more flexible to various conditioned portraits. Besides, we construct a high-quality wide-angle portrait dataset and design two metrics for quantitative evaluations. Considerable experiments verify the robustness and generalization of our method. We believe this work is of great practical value. 

\section*{Acknowledgement}
This research was supported in part by National Natural Science Foundation of China (NSFC, No.61872067, No.61720106004), in part by Research Programs of Science and Technology in Sichuan Province (No.2019YFH0016).

{\small
%input{cvpr_arxiv.bbl}
\bibliographystyle{ieee_fullname}
\bibliography{egbib}
}

\end{document}